\documentclass{article}

\PassOptionsToPackage{numbers}{natbib}  
\usepackage[final]{nips_2017}

\usepackage[utf8]{inputenc} 
\usepackage[T1]{fontenc}    
\usepackage{hyperref}       
\usepackage{url}            
\usepackage{booktabs}       
\usepackage{amsmath}
\usepackage{amssymb}
\usepackage{amsfonts}       
\usepackage{nicefrac}       
\usepackage{microtype}      
\usepackage{cleveref}
\usepackage{graphicx}
\usepackage{color}
\usepackage{subcaption}
\usepackage{xspace}
\usepackage{float}

\makeatletter
\def\NAT@spacechar{~}  
\makeatother
\graphicspath{{figure/}}

\newcommand{\plotwidth}{.43}

\title{Learning Hierarchical Information Flow\\with Recurrent Neural Modules}

\author{
  Danijar Hafner\rlap{\,\thanks{Work done during an internship with Google Brain.}} \\
  Google Brain \\
  \texttt{mail@danijar.com} \\
  \And
  Alex Irpan \\
  Google Brain \\
  \texttt{alexirpan@google.com} \\
  \AND
  \hspace*{-2.22em}James Davidson \\
  \hspace*{-2.22em}Google Brain \\
  \hspace*{-2.22em}\texttt{jcdavidson@google.com} \\
  \And
  \hspace*{-1.43em}Nicolas Heess \\
  \hspace*{-1.43em}Google DeepMind \\
  \hspace*{-1.43em}\texttt{heess@google.com} \\
}

\begin{document}

\maketitle

\begin{abstract}
We propose ThalNet, a deep learning model inspired by neocortical communication via the thalamus. Our model consists of recurrent neural modules that send features through a routing center, endowing the modules with the flexibility to share features over multiple time steps. We show that our model learns to route information hierarchically, processing input data by a chain of modules. We observe common architectures, such as feed forward neural networks and skip connections, emerging as special cases of our architecture, while novel connectivity patterns are learned for the text8 compression task. Our model outperforms standard recurrent neural networks on several sequential benchmarks.
\end{abstract}

\section{Introduction}
\label{introduction}

Deep learning models make use of modular building blocks such as fully connected layers, convolutional layers, and recurrent layers. Researchers often combine them in strictly layered or task-specific ways. Instead of prescribing this connectivity a priori, our method learns how to route information as part of learning to solve the task. We achieve this using recurrent modules that communicate via a routing center that is inspired by the thalamus.

Warren McCulloch and Walter Pitts invented the perceptron in 1943 as the first mathematical model of neural information processing~\cite{mcculloch1943perceptron}, laying the groundwork for modern research on artificial neural networks. Since then, researchers have continued looking for inspiration from neuroscience to identify new deep learning architectures~\cite{hawkins2006hierarchical,hinton2011transforming,kirkpatrick2017overcoming,zenke2017improved}.

While some of these efforts have been directed at learning biologically plausible mechanisms in an attempt to explain brain behavior, our interest is to achieve a flexible learning model. In the neocortex, communication between areas can be broadly classified into two pathways: Direct communication and communication via the thalamus~\cite{sherman2016thalamus}. In our model, we borrow this latter notion of a centralized routing system to connect specializing neural modules.

In our experiments, the presented model learns to form connection patterns that process input hierarchically, including skip connections as known from ResNet~\cite{he2015resnet}, Highway networks \cite{srivastava2015highway}, and DenseNet~\cite{huang2016densely} and feedback connections, which are known to both play an important role in the neocortex and improve deep learning~\cite{gilbert2007brain,lillicrap2016feedback}. The learned connectivity structure is adapted to the task, allowing the model to trade-off computational width and depth. In this paper, we study these properties with the goal of building an understanding of the interactions between recurrent neural modules.

Section~\ref{model} defines our computational model. We point out two critical design axes, which we explore experimentally in the supplementary material. In Section~\ref{performance} we compare the performance of our model on three sequential tasks, and show that it consistently outperforms multi-layer recurrent networks. In Section~\ref{connectivity}, we apply the best performing design to a language modeling task, where we observe that the model automatically learns hierarchical connectivity patterns.
\section{Thalamus Gated Recurrent Modules}
\label{model}

\begin{figure}
\centering
\begin{subfigure}[t]{.35\textwidth}
\centering
\includegraphics[width=\textwidth]{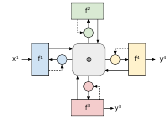}
\caption{Module $f^1$ receives the task input, $f^2$ can be used for side computation, $f^3$ is trained on an auxiliary task, and $f^4$ produces the output for the main task.}
\label{fig:model}
\end{subfigure}\hfill
\begin{subfigure}[t]{.6\textwidth}
\centering
\includegraphics[width=\textwidth]{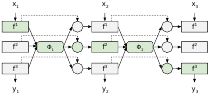}
\caption{Computation of 3 modules unrolled in time. One possible path of hierarchical information flow is highlighted in green. We show that our model learns hierarchical information flow, skip connections and feedback connections in Section~\ref{connectivity}.}
\label{fig:unrolled}
\end{subfigure}\hfill
\caption{Several modules share their learned features via a routing center. Dashed lines are used for dynamic reading only. We define both static and dynamic reading mechanisms in Section~\ref{reading}.}
\end{figure}

We find inspiration for our work in the neurological structure of the neocortex. Areas of the neocortex communicate via two principal pathways: The \emph{cortico-cortico-pathway} comprises direct connections between nuclei, and the \emph{cortico-thalamo-cortico} comprises connections relayed via the thalamus. Inspired by this second pathway, we develop a sequential deep learning model in which modules communicate via a routing center. We name the proposed model ThalNet.

\subsection{Model Definition}
\label{subsect:model-defn}
Our system comprises a tuple of computation modules $F=(f^1,\cdots,f^I)$ that route their respective features into a shared center vector $\Phi$. An example instance of our ThalNet model is shown in Figure~\ref{fig:model}. At every time step $t$, each module $f^i$ reads from the center vector via a context input $c^i_t$ and an optional task input $x^i_t$. The features $\phi^i_t=f^i(c^i_t, x^i_t)$ that each module produces are directed into the center $\Phi$.\footnote{In practice, we experiment with both feed forward and recurrent implementations of the modules $f^i$. For simplicity, we omit the hidden state used in recurrent modules in our notation.}\ Output modules additionally produce task output from their feature vector as a function $o^i(\phi^i)=y^i$.

All modules send their features to the routing center, where they are merged to a single feature vector $\Phi_t=m(\phi^1_t,\cdots,\phi^I_t)$. In our experiments, we simply implement $m$ as the concatenation of all $\phi^i$. At the next time step, the center vector $\Phi_t$ is then read selectively by each module using a reading mechanism to obtain the context input $c^i_{t+1}=r^i(\Phi_t,\phi^i_t)$.\footnote{The reading mechanism is conditioned on both $\Phi_t$ and $\phi^i_t$ separately as the merging does not preserve $\phi^i_t$ in the general case.}\ This reading mechanism allows modules to read individual features, allowing for complex and selective reuse of information between modules. The initial center vector $\Phi_0$ is the zero vector.

\pagebreak

In summary, ThalNet is governed by the following equations:
\begin{flalign}
\hspace{3.8em}\makebox[3.8em][l]{Module features:}    && \phi^i_t &= f^i(c^i_t,x^i_t) &\label{eq:model_1}\\
\hspace{3.8em}\makebox[3.8em][l]{Module output:}      && y^i_t &= o^i(\phi^i_t) &\label{eq:model_2}\\
\hspace{3.8em}\makebox[3.8em][l]{Center features:}    && \Phi_t &= m(\phi^1_t,\cdots,\phi^I_t) &\label{eq:model_3}\\
\hspace{3.8em}\makebox[3.8em][l]{Read context input:} && c^i_{t+1} &= r^i(\Phi_t,\phi^i_t)\label{eq:model_4}
\end{flalign}

The choice of input and output modules depends on the task at hand. In a simple scenario (e.g., single task), there is exactly one input module receiving task input, some number of side modules, and exactly one output module producing predictions. The output modules get trained using appropriate loss functions, with their gradients flowing backwards through the fully differentiable routing center into all modules.

Modules can operate in parallel as reads target the center vector from the previous time step. An unrolling of the multi-step process can be seen in Figure~\ref{fig:unrolled}. This figure illustrates the ability to arbitrarily route between modules between time steps This suggest a sequential nature of our model, even though application to static input is possible by allowing observing the input for multiple time steps.

We hypothesize that modules will use the center to route information through a chain of modules before producing the final output (see Section~\ref{connectivity}). For tasks that require producing an output at every time step, we repeat input frames to allow the model to process through multiple modules first, before producing an output. This is because communication between modules always spans a time step.\footnote{Please refer to \citet{graves2016adaptive} for a study of a similar approach.}

\subsection{Reading Mechanisms}
\label{reading}

\begin{figure}
\centering
\includegraphics[width=.6\textwidth]{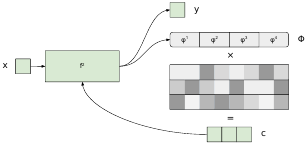}
\caption{The ThalNet model from the perspective of a single module. In this example, the module receives input $x^i$ and produces features to the center $\Phi$ and output $y^i$. Its context input $c^i$ is determined as a linear mapping of the center features from the previous time step. In practice, we apply weight normalization to encourage interpretable weight matrices (analyzed in Section~\ref{connectivity}).}
\label{fig:reading}
\end{figure}

We now discuss implementations of the reading mechanism $r^i(\Phi,\phi^i)$ and modules $f^i(c^i,x^i)$, as defined in Section~\ref{subsect:model-defn}. We draw a distinction between static and dynamic reading mechanisms for ThalNet. For static reading, $r^i(\Phi)$ is conditioned on independent parameters. For dynamic reading, $r^i(\Phi,\phi^i)$ is conditioned on the current corresponding module state, allowing the model to adapt its connectivity within a single sequence. We investigate the following reading mechanisms:

\begin{itemize}

\item \textbf{Linear Mapping.}\quad In its simplest form, static reading consists of a fully connected layer $r(\Phi,\cdot)=W\Phi$ with weights $W\in\mathbb{R}^{|c|\times|\Phi|}$ as illustrated in Figure~\ref{fig:reading}. This approach performs reasonably well, but can exhibit unstable learning dynamics and learns noisy weight matrices that are hard to interpret. Regularizing weights using L1 or L2 penalties does not help here since it can cause side modules to not get read from anymore.

\item \textbf{Weight Normalization.}\quad We found linear mappings with weight normalization~\citep{salimans2016weightnorm} parameterization to be effective. For this, the context input is computed as $r(\Phi,\cdot)=\beta\frac{W}{|W|}\Phi$ with scaling factor $\beta\in\mathbb{R},$ weights $W\in\mathbb{R}^{|c|\times|\Phi|},$ and the Euclidean matrix norm $|W|.$ Normalization results in interpretable weights since increasing one weight pushes other, less important, weights closer to zero, as demonstrated in Section~\ref{connectivity}.

\item \textbf{Fast Softmax.}\quad To achieve dynamic routing, we condition the reading weight matrix on the current module features $\phi^i$. This can be seen as a form of fast weights, providing a biologically plausible method for attention~\citep{ba2016fastweights,schmidhuber1992fastweights}. We then apply softmax normalization to the computed weights so that each element of the context is computed as a weighted average over center elements, rather than just a weighted sum. Specifically, $r(\Phi,\phi)_{(j)}=\big({e^{(W\phi+b)_{(j)}}}/{\sum_{k=1}^{|\Phi|}{e^{(W\phi+b)_{(jk)}}}}\big)\Phi$ with weights $W\in\mathbb{R}^{|\phi|\times|\Phi|\times|c|},$ and biases $b\in\mathbb{R}^{|\Phi|\times|c|}.$ While this allows for a different connectivity pattern at each time step, it introduces ${|\phi^i+1|\times|\Phi|\times|c^i|}$ learned parameters per module.

\item \textbf{Fast Gaussian.}\quad As a compact parameterization for dynamic routing, we consider choosing each context element as a Gaussian weighted average of $\Phi$, with only mean and variance vectors learned conditioned on $\phi^i$. The context input is computed as $r(\Phi,\phi)_{(j)}=f\big((1,2,\cdots,|\Phi|)|(W\phi+b)_{(j)},(U\phi+d)_{(j)}\big)\Phi$ with weights $W,U\in\mathbb{R}^{|c|\times|\phi|},$ biases $b,d\in\mathbb{R}^{|c|},$ and the Gaussian density function $f(x|\mu,\sigma^2)$. The density is evaluated for each index in $\Phi$ based on its distance from the mean. This reading mechanism only requires ${|\phi^i+1|\times 2\times|c^i|}$ parameters per module and thus makes dynamic reading more practical. 

\end{itemize}

Reading mechanisms could also select between modules on a high level, instead of individual feature elements. We do not explore this direction since it seems less biologically plausible. Moreover, we demonstrate that such knowledge about feature boundaries is not necessary, and hierarchical information flow emerges when using fine-grained routing (see Figure~\ref{fig:weights}). Theoretically, this also allows our model to perform a wider class of computations.

\section{Performance Comparison}
\label{performance}

We investigate the properties and performance of our model on several benchmark tasks. First, we compare reading mechanisms and module designs on a simple sequential task, to obtain a good configuration for the later experiments. Please refer to the supplementary material for the precise experiment description and results. We find that the weight normalized reading mechanism provides best performance and stability during training. We will use ThalNet models with four modules of configuration for all experiments in this section. To explore the performance of ThalNet, we now conduct experiments on three sequential tasks of increasing difficulty:

\begin{itemize}

\item \textbf{Sequential Permuted MNIST.}\quad We use images from the MNIST~\citep{lecun1998mnist} data set, the pixels of every image by a fixed random permutation, and show them to the model as a sequence of rows. The model outputs its prediction of the handwritten digit at the last time step, so that it must integrate and remember observed information from previous rows. This delayed prediction combined with the permutation of pixels makes the task harder than the static image classification task, with a multi-layer recurrent neural network achieving \textasciitilde 65\,\% test error. We use the standard split of 60,000 training images and 10,000 testing images.

\item \textbf{Sequential CIFAR-10.}\quad In a similar spirit, we use the CIFAR-10~\citep{cifar10} data set and feed images to the model row by row. We flatten the color channels of every row so that the model observes a vector of 96 elements at every time step. The classification is given after observing the last row of the image. This task is more difficult than the MNIST task, as the image show more complex and often ambiguous objects. The data set contains 50,000 training images and 10,000 testing images.

\item \textbf{Text8 Language Modeling.}\quad This text corpus consisting of the first $10^8$ bytes of the English Wikipedia is commonly used as a language modeling benchmark for sequential models. At every time step, the model observes one byte, usually corresponding to 1 character, encoded as a one-hot vector of length 256. The task it to predict the distribution of the next character in the sequence. Performance is measured in bits per character (BPC) computed as $-\frac{1}{N}\sum^N_{i=1}{\log_2p(x_i)}$. Following \citet{cooijmans2016recbatchnorm}, we train on the first 90\% and evaluate performance on the following 5\% of the corpus.

\end{itemize}

For the two image classification tasks, we compare variations of our model to a stacked Gated Recurrent Unit (GRU)~\cite{cho2014gru} network of 4 layers as baseline. The variations we compare are different choices of feed-forward layers and GRU layers for implementing the modules $f^i(c^i,x^i)$: We test with two fully connected layers (FF), a GRU layer (GRU), fully connected followed by GRU (FF-GRU), GRU followed by fully connected (GRU-FF), and a GRU sandwiched between fully connected layers (FF-GRU-FF).\footnote{Note that the modules require some amount of local structure to allow them to specialize. Implementing the modules as a single fully connected layer recovers a standard recurrent neural network with one large layer.}\ For all models, we pick the largest layer sizes such that the number of parameters does not exceed 50,000. Training is performed for 100 epochs on batches of size 50 using RMSProp~\cite{tieleman2012rmsprop} with a learning rate of $10^{-3}$.

For language modeling, we simulate ThalNet for 2 steps per token, as described in Section~\ref{model} to allow the output module to read information about the current input before making its prediction. Note that on this task, our model uses only half of its capacity directly, since its side modules can only integrate longer-term dependencies from previous time steps. We run the baseline once without extra steps and once with 2 steps per token, allowing it to apply its full capacity once and twice on each token, respectively. This makes the comparison a bit difficult, but only by favouring the baseline. This suggests that architectural modifications, such as explicit skip-connections between modules, could further improve performance.

\begin{figure}[t]
\centering
\begin{subfigure}[t]{.33\textwidth}
\centering
\includegraphics[width=\textwidth]{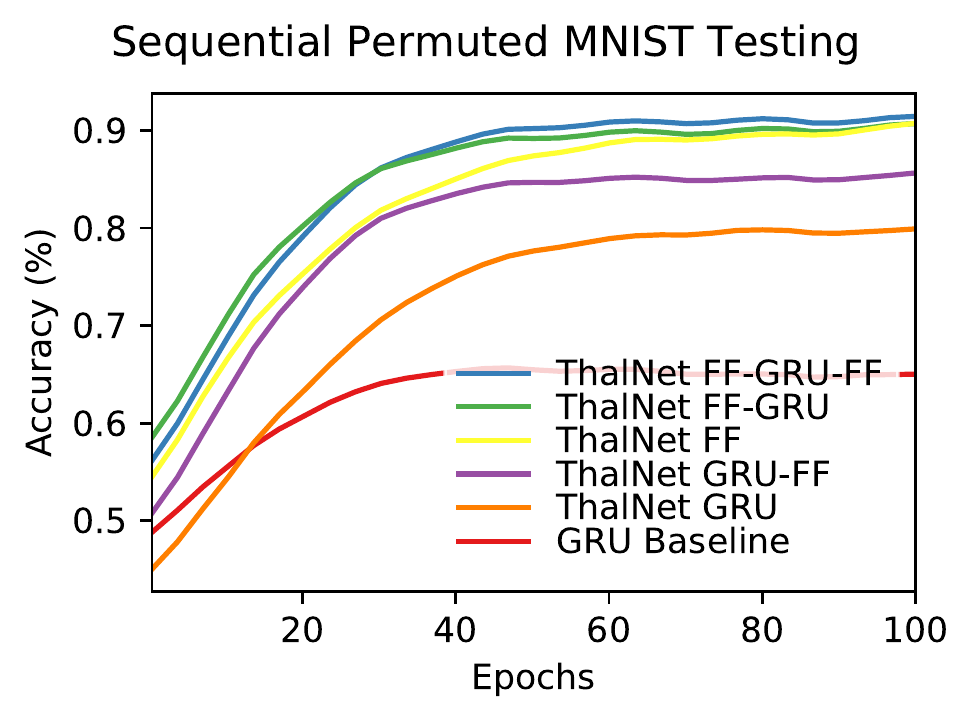}
\label{fig:perm-mnist-row-eval}
\end{subfigure}%
\begin{subfigure}[t]{.33\textwidth}
\centering
\includegraphics[width=\textwidth]{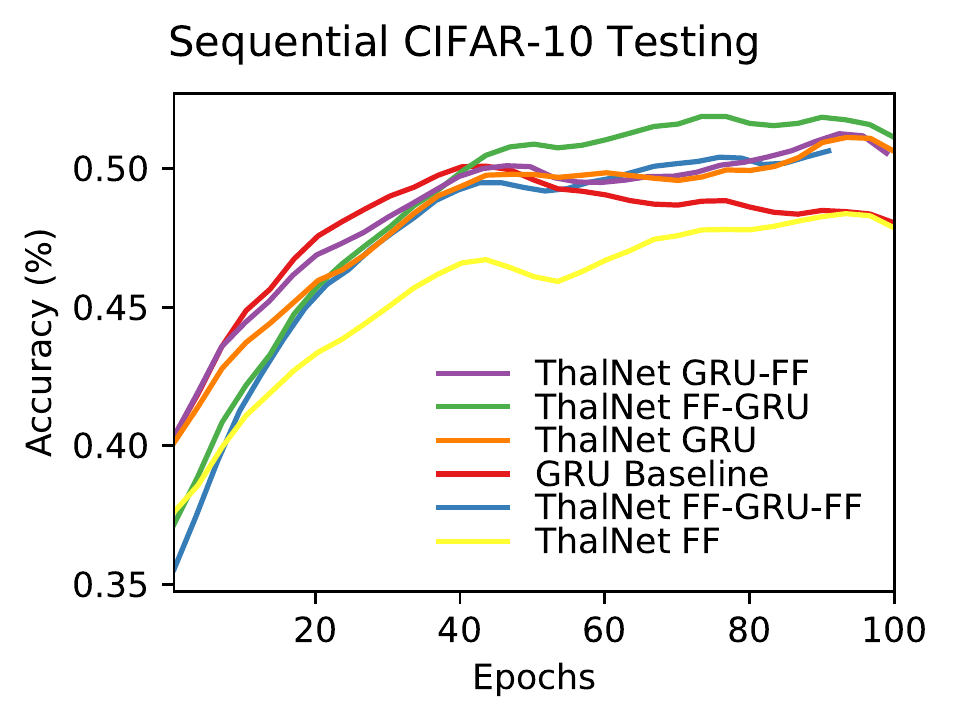}
\label{fig:cifar-eval}
\end{subfigure}%
\begin{subfigure}[t]{.33\textwidth}
\centering
\includegraphics[width=\textwidth]{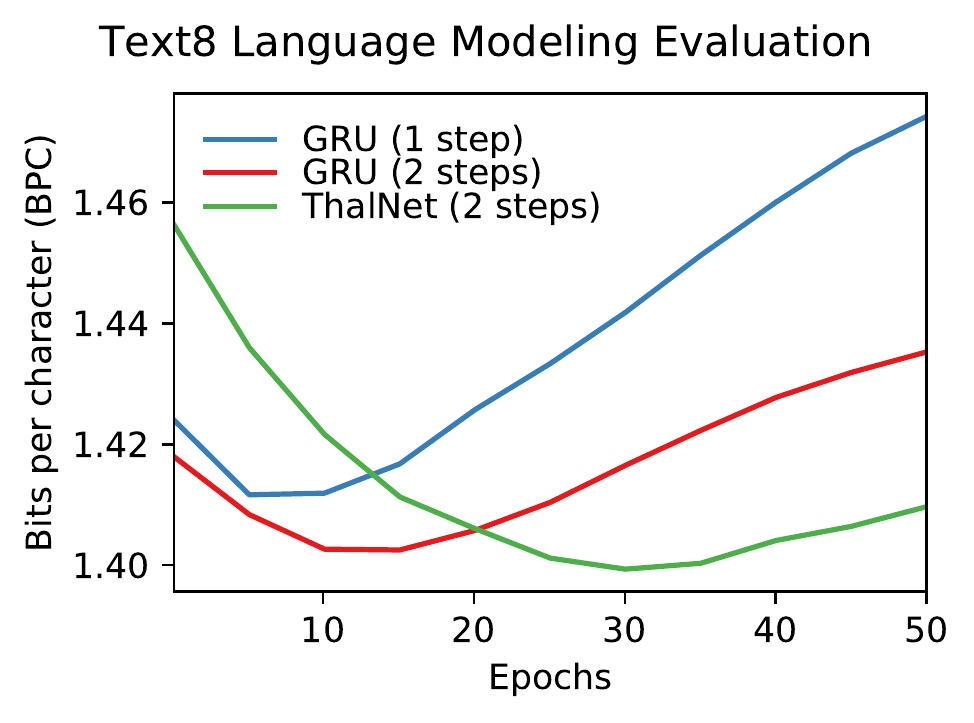}
\label{fig:text8-eval}
\end{subfigure}%
\vspace{-2ex}
\begin{subfigure}[t]{.33\textwidth}
\centering
\includegraphics[width=\textwidth]{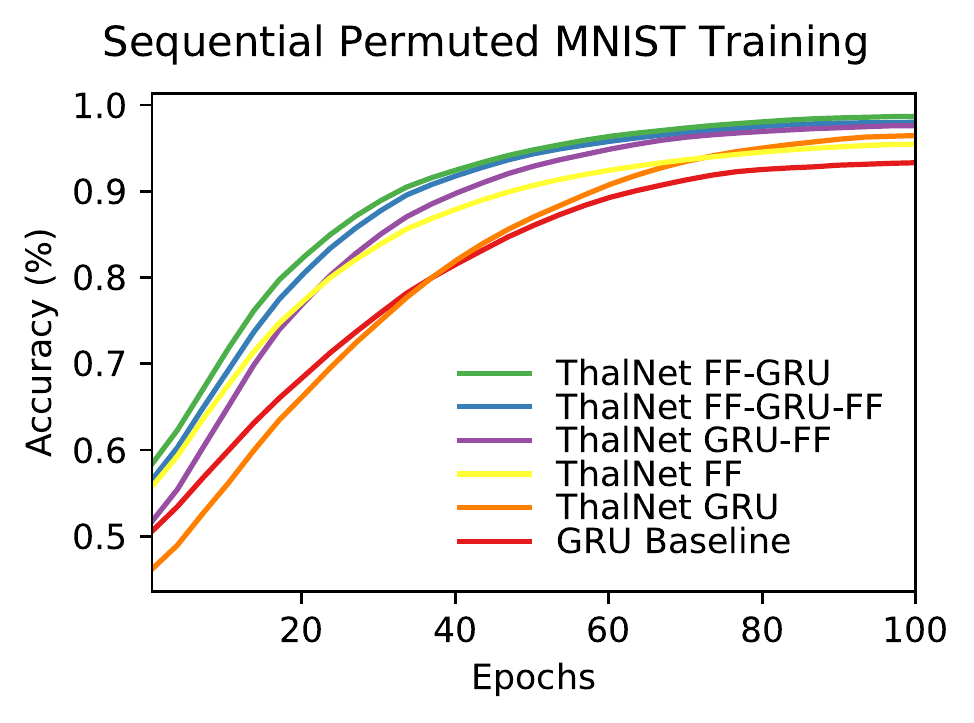}
\label{fig:perm-mnist-row-train}
\end{subfigure}%
\begin{subfigure}[t]{.33\textwidth}
\centering
\includegraphics[width=\textwidth]{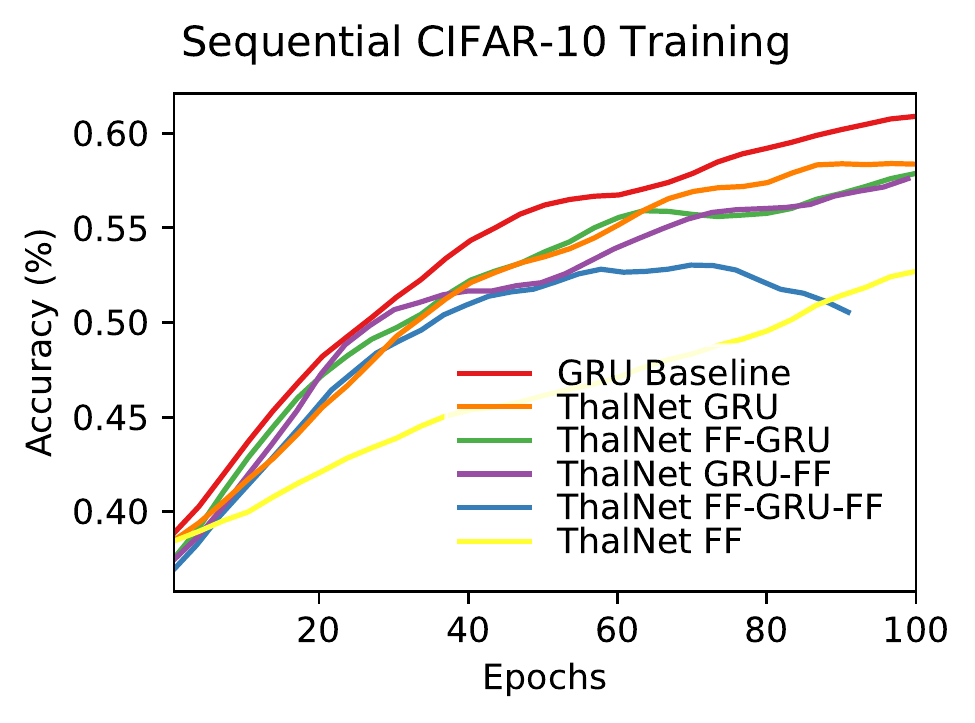}
\label{fig:cifar-train}
\end{subfigure}%
\begin{subfigure}[t]{.33\textwidth}
\centering
\includegraphics[width=\textwidth]{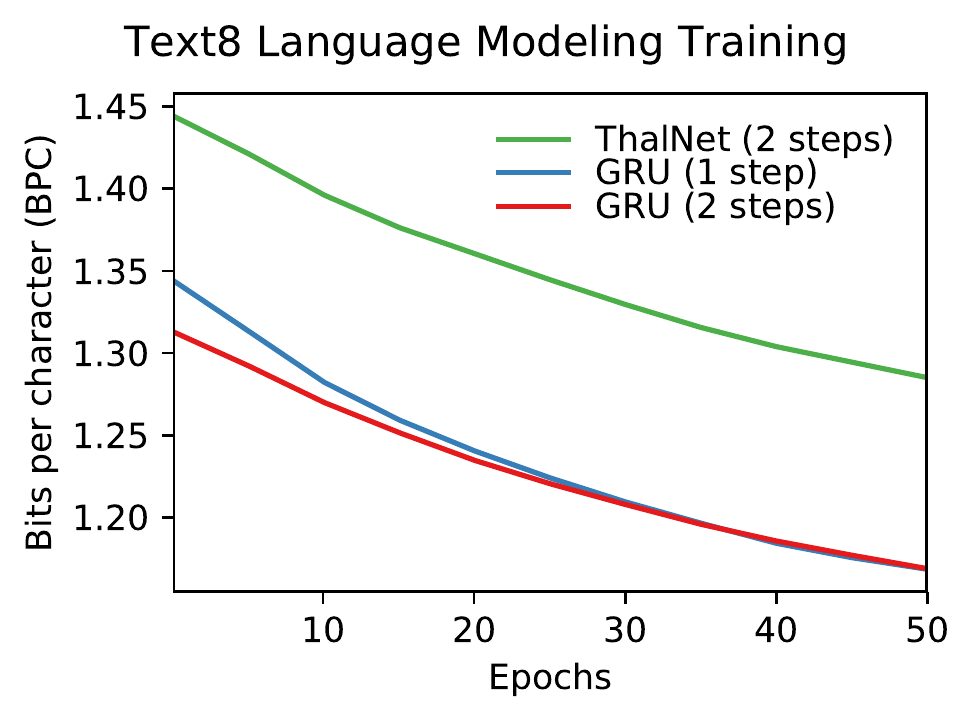}
\label{fig:text8-train}
\end{subfigure}%
\caption{Performance on the permuted sequential MNIST, sequential CIFAR, and text8 language modeling tasks. The stacked GRU baseline reaches higher training accuracy on CIFAR, but fails to generalize well. On both tasks, ThalNet clearly outperforms the baseline in testing accuracy. On CIFAR, we see how recurrency within the modules speeds up training. The same pattern is shows for the text8 experiment, where ThalNet using 12M parameters matches the performance of the baseline with 14M parameters. The step number 1 or 2 refers to repeated inputs as discussed in Section~\ref{model}. We had to smooth the graphs using a running average since the models were evaluated on testing batches on a rolling basis.}
\label{fig:performance}
\end{figure}

The Text8 task requires larger models. We train ThalNet with 4 modules of a size 400 feed forward layer and a size 600 GRU layer each, totaling in 12 million model parameters. We compare to a standard baseline in language modeling, a single GRU with 2000 units, totaling in 14 million parameters. We train on batches of 100 sequences, each containing 200 bytes, using the Adam optimizer~\cite{kingma2014adam} with a default learning rate of $10^{-3}$. We scale down gradients exceeding a norm of $1$. Results for 50 epochs of training are shown in Figure~\ref{fig:performance}. The training took about 8 days for ThalNet with 2 steps per token, 6 days for the baseline with 2 steps per token, and 3 days for the baseline without extra steps.

Figure~\ref{fig:performance} shows the training and testing and training curves for the three tasks described in this section. ThalNet outperforms standard GRU networks in all three tasks. Interestingly, ThalNet experiences a much smaller gap between training and testing performance than our baseline -- a trend we observed across all experimental results.

On the Text8 task, ThalNet scores 1.39 BPC using 12M parameters, while our GRU baseline scores 1.41 BPC using 14M parameters (lower is better). Our model thus slightly improves on the baseline while using fewer parameters. This result places ThalNet in between the baseline and regularization methods designed for language modeling, which can also be applied to our model. The baseline performance is consistent with published results of LSTMs with similar number of parameters~\citep{krueger2016zoneout}.

We hypothesize the information bottleneck at the reading mechanism acting as an implicit regularizer that encourages generalization. Compared to using one large RNN that has a lot of freedom of modeling the input-output mapping, ThalNet imposes local structure to how the input-output mapping can be implemented. In particular, it encourages the model to decompose into several modules that have stronger intra-connectivity than extra-connectivity. Thus, to some extend every module needs to learn a self-contained computation.
\section{Hierarchical Connectivity Patterns}
\label{connectivity}

Using its routing center, our model is able to learn its structure as part of learning to solve the task. In this section, we explore the emergent connectivity patterns. We show that our model learns to route features in hierarchical ways as hypothesized, including skip connections and feedback connections. For this purpose, we choose the text8 corpus, a medium-scale language modeling benchmark consisting of the first $10^8$ bytes of Wikipedia, preprocessed for the Hutter~Prize~\cite{mahoney2011text8}. The model observes one one-hot encoded byte per time step, and is trained to predict its future input at the next time step.

We use comparably small models to be able to run experiments quickly, comparing ThalNet models of 4 FF-GRU-FF modules with layer sizes 50, 100, 50 and 50, 200, 50. Both experiments use weight normalized reading. Our focus here is on exploring learned connectivity patterns. We show competitive results on the task using larger models in Section~\ref{performance}.

We simulate two sub time steps to allow for the output module to receive information of the current input frame as discussed in Section~\ref{model}. Models are trained for 50 epochs on batches of size 10 containing sequences of length 50 using RMSProp with a learning rate of $10^{-3}$. In general, we observe different random seeds converging to similar connectivity patterns with recurring elements.

\begin{figure}
\centering
\begin{subfigure}[t]{.31\textwidth}
\centering
\includegraphics[width=\textwidth]{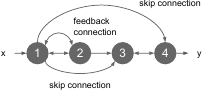}\vspace{1ex}
\includegraphics[width=\textwidth]{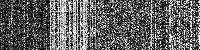}\vspace{0.5ex}
\includegraphics[width=\textwidth]{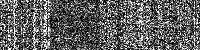}\vspace{0.5ex}
\includegraphics[width=\textwidth]{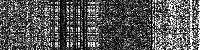}\vspace{0.5ex}
\includegraphics[width=\textwidth]{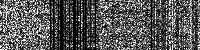}
\caption{Weight Normalization}
\label{fig:weights-weight-norm}
\end{subfigure}\hfill%
\begin{subfigure}[t]{.31\textwidth}
\centering
\includegraphics[width=\textwidth]{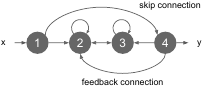}\vspace{1ex}
\includegraphics[width=\textwidth]{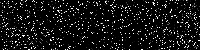}\vspace{0.5ex}
\includegraphics[width=\textwidth]{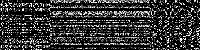}\vspace{0.5ex}
\includegraphics[width=\textwidth]{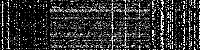}\vspace{0.5ex}
\includegraphics[width=\textwidth]{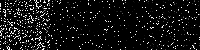}
\caption{Fast Softmax}
\label{fig:weights-fast-softmax}
\end{subfigure}\hfill%
\begin{subfigure}[t]{.31\textwidth}
\centering
\vspace{1.3ex}
\includegraphics[width=\textwidth]{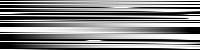}\vspace{0.5ex}
\includegraphics[width=\textwidth]{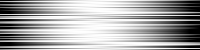}\vspace{0.5ex}
\includegraphics[width=\textwidth]{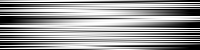}\vspace{0.5ex}
\includegraphics[width=\textwidth]{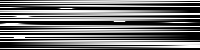}
\caption{Fast Gaussian}
\label{fig:weights-fast-gaussian}
\end{subfigure}%
\caption{Reading weights learned by different reading mechanisms with 4 modules on the text8 language modeling task, alongside manually deducted connectivity graphs. We plot the weight matrices that produce the context inputs to the four modules, top to bottom. The top images show focus of the input modules, followed by side modules, and output modules at the bottom. Each pixel row gets multiplied with the center vector $\Phi$ to produce one scalar element of the context input $c^i$. We visualize the magnitude of weights between the 5\,\% to the 95\,\% percentile. We do not include the connectivity graph for Fast Gaussian reading as its reading weights are not clearly structured.}
\label{fig:weights}
\end{figure}

\subsection{Trained Reading Weights}

Figure~\ref{fig:weights} shows trained reading weights for various reading mechanisms, along with their connectivity graphs that were manually deduced.\footnote{Developing formal measurements for this deduction process seems beneficial in the future.}\ Each image represents a reading weight matrix for the modules 1 to 4 (top to bottom). Each pixel row shows the weight factors that get multiplied with $\Phi$ to produce a single element of the context vector of that module. The weight matrices thus has dimensions of $|\Phi|\times|c^i|$. White pixels represent large magnitudes, suggesting focus on features at those positions.

The weight matrices of weight normalized reading clearly resemble the boundaries of the four concatenated module features $\phi^1,\cdots,\phi^4$ in the center vector $\Phi$, even though the model has no notion of the origin and ordering of elements in the center vector.

A similar structure emerges with fast softmax reading. These weight matrices are sparser than the weights from weight normalization. Over the course of a sequence, we observe some weights staying constant while others change their magnitudes at each time step. This suggests that optimal connectivity might include both static and dynamic elements. However, this reading mechanism leads to less stable training. This problem could potentially alleviated by normalizing the fast weight matrix.

With fast Gaussian reading, we see that the distributions occasionally tighten on specific features in the first and last modules, the modules that receive input and emit output. The other modules learn large variance parameters, effectively spanning all center features. This could potentially be addressed by reading using mixtures of Gaussians for each context element instead. We generally find that weight normalized and fast softmax reading select features with in a more targeted way.

\subsection{Commonly Learned Structures}

The top row in Figure~\ref{fig:weights} shows manually deducted connectivity graphs between modules. Arrows represent the main direction of information flow in the model. For example, the two incoming arrows to module 4 in Figure~\ref{fig:weights-weight-norm} indicate that module 4 mainly attends to features produced by modules 1 and 3. We infer the connections from the larger weight magnitudes in the first and third quarters of the reading weights for module 4 (bottom row).

A typical pattern that emerges during the experiments can be seen in the connectivity graphs of both weight normalized and fast softmax reading (Figures~\ref{fig:weights-weight-norm}~and~\ref{fig:weights-fast-softmax}). Namely, the output module reads features directly from the input module. This direction connection is established early on during training, likely because this is the most direct gradient path from output to input. Later on, the side modules develop useful features to support the input and output modules.

In another pattern, one module reads from all other modules and combines their information. In Figure~\ref{fig:weights-fast-softmax}, module 2 takes this role, reading from modules 1, 3, 4, and distributing these features via the input module. In additional experiments with more than four modules, we observed this pattern to emerge predominantly. This connection pattern provides a more efficient way of information sharing than cross-connecting all modules.

Both connectivity graphs in Figure~\ref{fig:weights} include hierarchical computation paths through the modules. They include learn skip connections, which are known to improve gradient flow from popular models such as ResNet~\cite{he2015resnet}, Highway networks~\cite{srivastava2015highway}, and DenseNet~\cite{huang2016densely}. Furthermore, the connectivity graphs contain backward connections, creating feedback loops over two or more modules. Feedback connections are known to play a critical role in the neocortex, which inspired our work~\cite{gilbert2007brain}.

\section{Related Work}
\label{related}

We describe a recurrent mixture of experts model, that learns to dynamically pass information between the modules. Related approaches can be found in various recurrent and multi-task methods as outlined in this section.

\textbf{Modular Neural Networks.}\quad ThalNet consists of several recurrent modules that interact and exploit each other. Modularity is a common property of existing neural models. \cite{devin2016learning} learn a matrix of tasks and robot bodies to improve both multitask and transfer learning. \cite{andreas2016neural} learn modules modules specific to objects present in the scene, which are selected by an object classifier. These approaches specify modules corresponding to a specific task or variable manually. In contrast, our model automatically discovers and exploits the inherent modularity of the task and does not require a one-to-one correspondence of modules to task variables.

The Column Bundle model~\cite{pham2017columnbundle} consists of a central column and several mini-columns around it. While not applied to temporal data, we observe a structural similarity between our modules and the mini-columns, in the case where weights are shared among layers of the mini-columns, which the authors mention as a possibility.

\textbf{Learned Computation Paths.}\quad We learn the connectivity between modules alongside the task. There are various methods in the multi-task context that also connectivity between modules. \citet{fernando2017pathnet} learn paths through multiple layers of experts using an evolutionary approach. \citet{rusu2016progressive} learn adapter connections to connect to fixed previously trained experts and exploit their information. These approaches focus on feed-forward architectures. The recurrency in our approach allows for complex and flexible computational paths. Moreover, we learn interpretable weight matrices that can be examined directly without performing costly sensitivity analysis.

The Neural Programmer Interpreted presented by \citet{reed2015neural} is related to our dynamic gating mechanisms. In their work, a network recursively calls itself in a parameterized way to perform tree-shaped computations. In comparison, our model allows for parallel computation between modules and for unrestricted connectivity patterns between modules.

\textbf{Memory Augmented RNNs.}\quad The center vector in our model can be interpreted as an external memory, with multiple recurrent controllers operating on it. Preceding work proposes recurrent neural networks operating on external memory structures. The Neural Turing Machine proposed by \citet{graves2014neural}, and follow-up work~\cite{graves2016hybrid}, investigate differentiable ways to address a memory for reading and writing. In the ThalNet model, we use multiple recurrent controllers accessing the center vector. Moreover, our center vector is recomputed at each time step, and thus should not be confused with a persistent memory as is typical for model with external memory.
\section{Conclusion}
\label{conclusion}

We presented ThalNet, a recurrent modular framework that learns to pass information between neural modules in a hierarchical way. Experiments on sequential and permuted variants of MNIST and CIFAR-10 are a promising sign of the viability of this approach. In these experiments, ThalNet learns novel connectivity patterns that include hierarchical paths, skip connections, and feedback connections.

In our current implementation, we assume the center features to be a vector. Introducing a matrix shape for the center features would open up ways to integrate convolutional modules and similarity-based attention mechanisms for reading from the center. While matrix shaped features are easily interpretable for visual input, it is less clear how this structure will be leveraged for other modalities.

A further direction of future work is to apply our paradigm to tasks with multiple modalities for inputs and outputs. It seems natural to either have a separate input module for each modality, or to have multiple output modules that can all share information through the center. We believe this could be used to hint specialization into specific patterns and create more controllable connectivity patterns between modules. Similarly, we an interesting direction is to explore the proposed model can be leveraged to learn and remember a sequence of tasks.

We believe modular computation in neural networks will become more important as researchers approach more complex tasks and employ deep learning to rich, multi-modal domains. Our work provides a step in the direction of automatically organizing neural modules that leverage each other in order to solve a wide range of tasks in a complex world.

\clearpage
\bibliography{references}

\clearpage
\appendix
\section*{\LARGE\centering Supplementary Material for Learning Hierarchical Information Flow with Recurrent Neural Modules}\vspace{5ex}

\section{Module Designs and Reading Mechanisms}

\begin{figure}[H]
\centering
\begin{subfigure}{\plotwidth\textwidth}
\centering
\includegraphics[width=\textwidth]{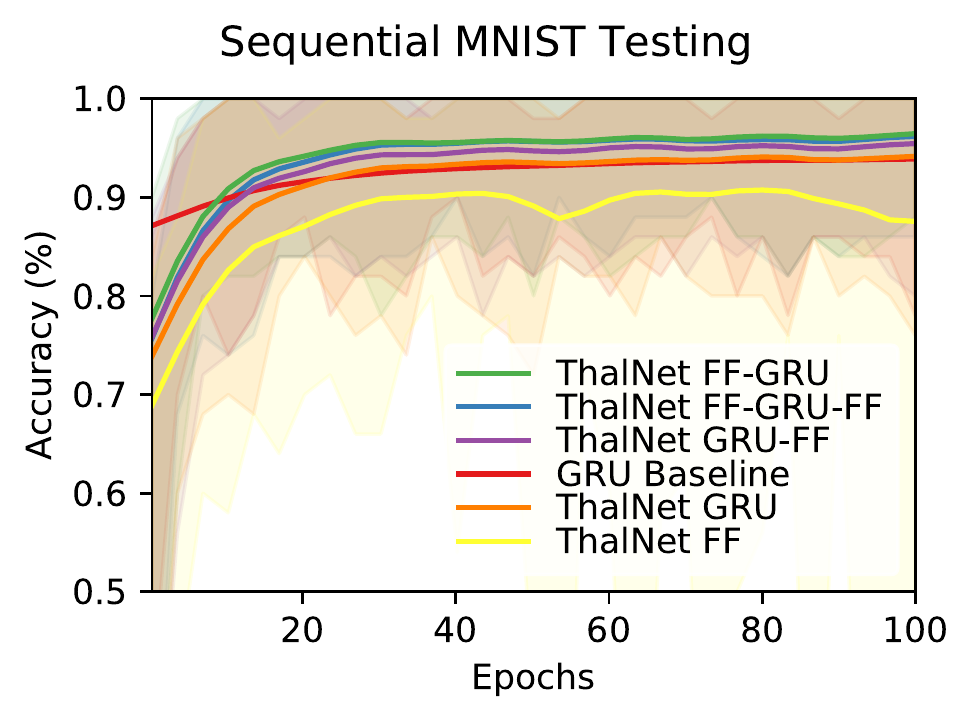}
    \caption{Module designs}
\label{fig:mnist-module}
\end{subfigure}\quad%
\begin{subfigure}{\plotwidth\textwidth}
\centering
\includegraphics[width=\textwidth]{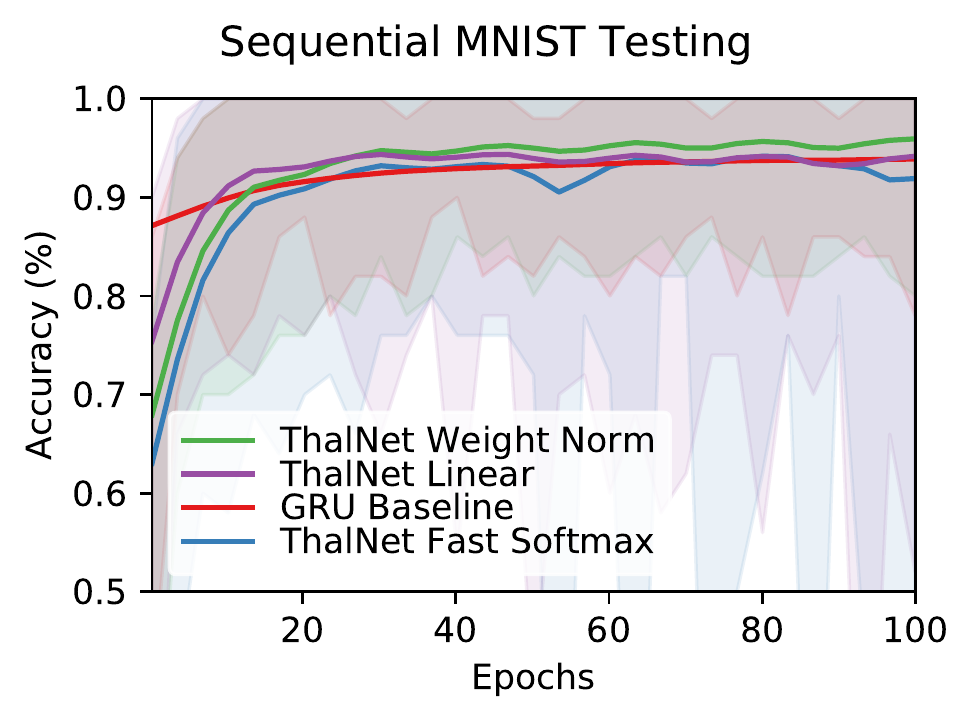}
    \caption{Reading mechanisms}
\label{fig:mnist-reading}
\end{subfigure}%
\caption{Test performance on the Sequential MNIST task grouped by module design (left) and reading mechanism (right). Plots show the top, median, and bottom accuracy over the other design choices. Recurrent modules train faster than pure fully connected modules and weight normalized reading is both stable and performs best. FF-GRU-FF modules perform similarly to FF-GRU while limiting the size of the center.}
\label{fig:mnist-graphs}
\end{figure}

We use a sequential variant of MNIST~\citep{lecun1998mnist} to compare the reading mechanisms described in Section~\ref{reading}, along with implementations of the module function. In Sequential MNIST, the model observes handwritten digits of $28\times 28$ pixels from top to bottom, one row per time step. The prediction is given at the last time step, so that the model has to integrate and remember observed information over the sequence. This makes the task more challenging than in the static setting with a multi-layer recurrent network achieving \textasciitilde 7\,\% error on this task.

To implement the modules $f^i(c^i,x^i)$ we test various combinations of fully connected and recurrent layers of Gated Recurrent Units (GRU)~\cite{cho2014gru}. Modules require some amount of local structure to allow them to specialize.\footnote{Implementing the modules as a single fully connected layer recovers a standard recurrent neural network with one large layer.}\ We test with two fully connected layers (FF), a GRU layer (GRU), fully connected followed by GRU (FF-GRU), GRU followed by fully connected (GRU-FF), and a GRU sandwiched between fully connected layers (FF-GRU-FF). In addition, we compare performance to a stacked GRU baseline with 4 layers. For all models, we pick the largest layer sizes such that the number of parameters does not exceed 50,000.

We train for 100 epochs on batches of size 50 using RMSProp~\cite{tieleman2012rmsprop} with a learning rate of $10^{-3}$. Figure~\ref{fig:mnist-graphs} shows the test accuracy of module designs and reading mechanisms. ThalNet outperforms the stacked GRU baseline in most configurations. We assume that the structure imposed by our model acts as a regularizer. We perform a further performance comparison in Section~\ref{performance}.

Results for module designs are shown in Figure~\ref{fig:mnist-module} in the appendix. We observe a benefit of recurrent modules as they exhibit faster and more stable training than fully connected modules. This could be explained by the fact that pure fully connected modules have to learn to use the routing center to store information over time, which is a long feedback loop. Having a fully connected layer before the recurrent layer also significantly improves performance. A fully connected layer after the GRU let us produce compact feature vectors $\phi^i$ that scale better to large modules, although we find FF-GRU to be beneficial in later experiments (Section~\ref{performance}).

Results for the reading mechanisms area shown in Figure~\ref{fig:mnist-reading}. The reading mechanism only has a small impact on the model performance. We find weight normalized reading to yield more stable performance than linear or fast softmax reading. For all further experiments, we use weight normalized reading due to both its stability and predictive performance. We do not include results for fast Gaussian reading here, as it performed below the performance range of the other methods.

\section{Interpretation as Recurrent Mixture of Experts}
\label{mixture}

ThalNet can route information from the input to the output over multiple time steps. This enables it to trade off shallow and deep computation paths. To understand this, we view ThalNet as a smooth mixture of experts model~\cite{jacobs1991task}, where the modules $F=(f^1,\cdots,f^I)$ are the recurrent experts. Each module outputs its features to the center vector $\Phi_t$. A linear combination of $\Phi_t$ is read at the next time step, which effectively performs a mixing of expert outputs. Compared to the recurrent mixture of experts model presented by \citet{shazeer2017outrageously}, our model can recurrently route information through the mixture of multiple times, increasing the number of mixture compounds.

To highlight two extreme cases, the modules could read from identical locations in the center. In this case, the model does a wide and shallow computation over 1 time step, analogous to \citet{graves2016adaptive}. In the other extreme, each module reads from a different module, recovering a hierarchy of recurrent layers. This gives a deep but narrow computation stretched over multiple time steps. In between, there exist a spectrum of complex patterns of information flow with differing and dynamic computation depths. This is comparable to DenseNet~\cite{huang2016densely}, which also blends information from paths of different computational depth, although in a purely feed-forward model.

Using state-less modules, our model could still leverage the recurrence between the modules and the center to store information over time. However, this bounds the number of distinct computation steps that ThalNet could apply to an input. Using recurrent modules, the computation steps can change over time, increasing the flexibility of the model. Recurrent modules give a stronger prior for using feedback and shows improved performance in our experiments.

\section{Comparison to Long Short-Term Memory}

When viewing the Equations~\ref{eq:model_1}\,--\,\ref{eq:model_4} in the model definition (Section~\ref{model}), one might think how our model compares to Long Short-Term Memory (LSTM)~\cite{hochreiter1997lstm}. However, there exists only a limited similarity between the two models. Empirically, we observed that LSTMs performed similarly to our GRU baselines when given the same parameter budget.

LSTM's context vector $c_t$ is processed element-wise, while ThalNet's routing center cross-connects modules. LSTM's hidden output $h_t$ is a better candidate for comparison with ThalNet's center features $\Phi$, which allows us to relate the recurrent weight matrix of an LSTM layer to the linear version of our reading mechanism.

We could relate each ThalNet module to a set of multiple LSTM units. However, LSTM units perform separate scalar computations, while our modules can learn complex interactions between multiple features at each time step. Alternatively, we could see LSTM units as very small ThalNet modules, reading exactly four context elements each, namely for the input and the three gates. However, the computational capacity and local structure of individual LSTM units is not comparable to that of the ThalNet modules used in our work.

\end{document}